\documentclass[conference]{IEEEtran}
\usepackage{times}

\usepackage[numbers]{natbib}
\usepackage{multicol}
\usepackage{graphicx}
\usepackage[normalem]{ulem}
\let\labelindent\relax
\usepackage{lipsum,graphicx,subcaption}
\usepackage[hyphens]{url}

\usepackage{enumitem}
\usepackage{amssymb}
\newlist{todolist}{itemize}{2}
\setlist[todolist]{label=$\square$}
\usepackage{amsmath}
\usepackage{amssymb}
\usepackage{gensymb}
\usepackage{algorithm}
\usepackage[noend]{algpseudocode}
\usepackage{caption}
\usepackage{subcaption}

\usepackage{hyperref}
\hypersetup{
    colorlinks=true,
    linkcolor=blue,
    filecolor=magenta,      
    urlcolor=cyan,
}

\usepackage{color}
\usepackage{xcolor}

\begin{document}

\title{Steps Towards Best Practices For Robot Videos}
\author{Eric Rosen, Stefanie Tellex, Geroge Konidaris \\
  Department of Computer Science\\
  Brown University\\
  Providence, RI \\ 
  er35@cs.brown.edu, stefie10@cs.brown.edu, gdk@cs.brown.edu }



%

\maketitle

\begin{abstract} 

There are unwritten guidelines for how to make robot videos that researchers learn from their advisors and pass onto their students. We believe that it is important for the community to collaboratively discuss and develop a standard set of best practices when making robot. We suggest a starting set of maxims for best robot video practices, and highlight positive examples from the community and negative examples only from videos made by the authors of this article. In addition, we offer a checklist that we hope can act as an document that can be given to robotic researchers to inform them of how to make robot videos that truthfully characterize what a robot can and can not do. We consider this a first draft, and are looking for feedback from the community as we refine and grow our maxims and checklist.


\end{abstract}

\IEEEpeerreviewmaketitle

\section{Introduction}
There are several types of research artifacts that the robotics community values: Research articles and papers, simulations, software and robot hardware are some of the primary ones. Robot videos are useful both as tools for story telling,  as well as as documenting robot research being applied in the real-world. Nick Roy says that the essence of robot research is to ``make the robot do something that it could not do before, and explain why''. Indeed, real-world robot challenges such as the Amazon Picking Challenge \cite{correll2016analysis,hernandez2016team}, RoboCup \cite{kitano1997robocup} and the DARPA challenge \cite{krotkov2017darpa} have spurred some of the most influential work in developing real-world applications of robotics research in the recent decade. Robot videos serve a uniquely distinct role of showing the viewer what the robot can do that it could not do before in the real world.

Despite the important role robot videos serve as a research artifact within the community, there has been no proposed standard set of best practices for making robot videos. In this article, we suggest a set of maxims for producing valuable robot videos for the research community: \textbf{make clear film edits}, \textbf{record all trials}, \textbf{display film rate}, and \textbf{highlight supporting systems}. We also discuss how various examples from the community already demonstrate these maxims, and offer a checklist based on these maxims that we hope act as a set of best practices for robot videos. We recognize that our set of maxims are not complete, and look forward to engaging with the community to further discuss and expand on a set of best practices for robot videos.

\section{Maxims for Robot Videos}
In this section, we offer a set of maxims (the title of each of subsection) for producing robot videos as valuable research artifacts for documenting what a robot can and can not do. We believe it is important for robot videos to accurately and transparently characterize the robustness, reliability, and completeness of robot systems capabilities. We offer advice on how to generally best follow the maxims, and discuss references to robot videos from the community that we find to be positive examples of each maxim and appropriate negative examples produced by the authors and collaborators.

\subsection{Clear Film Edits}
\label{sec:clearfilmedits}
Film editing is the processes of applying post-production techniques on video footage, and serves a variety of technical and artistic purposes when making robot videos. Technically, film editing is useful for purposes like simultaneously displaying and splicing together multiple camera views so as to inform the user what the robot is able to perceive as well as provide the viewer with more environmental context than can be achieved by a single camera perspective due to occlusion and limited sensing range. Artistically, film edits are useful for conveying a concise and compelling story narrative that represents what the robot is now able to do, by means such as removing footage where the robot does not move or performs a long, repetitive behavior, to combining video footage taken from different trials to demonstrate best-case scenario of the robot completing an entire task in a single attempt. We believe that film editing provides incredibly useful techniques for improving viewer comprehension and enjoyment, but recognize that it is important to make clear within a robot video how film edits are being used when it is not clear in order to best characterize what a robot can and can not do.

Long take footage is useful in robot videos for demonstrating how reliable the robot system is. One useful film editing technique with long take footage is to display synchronized footage from a single trial captured from multiple camera perspectives. For example, in the video made by \citet{mousavian20196}, they display continuous footage of a robot generating grasp points on objects and then performing a pick and place task over multiple trials (video at \cite{edit1video}). To give the viewer better environmental context, the video contains synchronized footage from a camera mounted on the end effector of the robot and from a third-party camera sitting statically next to the objects. The one-take footage plays over multiple trials, giving the viewer a nuanced understanding of how the objects are configured and what the robot egotistically perceives. 
Another example of clear film editing in a robot video is from \citet{kayacan2018embedded}, where they recorded footage of a robot autonomously counting corn stands while driving near farm crops (video at \cite{edit2video}). In the video, they display footage both from a third-party camera as well as camera footage captured from the forward on board camera, which they clearly label in their video. Clearly splicing together synchronised camera footage captured from a single trial is a great way of providing the viewer with contextual and environmental awareness for what the robot can do.

We also offer an example made by collaborators of this paper (\cite{bollini2013interpreting}) that uses film editing to artistically tell a story narrative for what the robot can do, but does not clearly indicate what film edits are made. In the video, a robot is making cookies, and there are many different camera views edited back-to-back (video at \cite{badeditvideo}). It is not clear if the footage taken comes from a single trial or multiple trials, and there is footage that is removed (stirring, etc.) without any mention regarding how much footage was removed or how many trials it took to get all the shots. While \cite{bollini2013interpreting} successfully uses film editing techniques to tell a compelling story of what the robot can now do, it does not make clear what is movie magic (artistic film editing) and what is robot magic (research contribution). We suggest that robot videos explicitly make clear when footage is taken from different trials and edited together, where camera footage is captured from in the scene, and when robot footage is removed. Robot videos should give the viewer an accurate sense of how long the trial took, how reliable and robust the system is, and whether the video was done in one trial.

\subsection{Record All Trials}
Robot experiments often require many trials and extended interaction with the environment. Robot videos should highlight when a robot performs a task successfully in order to give the viewer a sense of what the robot can do that it could not do before. However, it is also important and useful for the viewer to be able to get a sense of the system in both success and failure modes. Transparency about the limitations and problems of systems is important because that is where new research directions are born. Recording all robot trials is useful for providing rich documentation on how the robot system operates, and (if possible) it is best to preserve all footage so that it can accessed along with the shorter video.

Videos of robots failing are both popular among the public as well as valuable research artifacts in the robotics community. Footage of robots falling over at the DARPA robotics challenge (video at \cite{darpafail}) are popular because they are humorous as well as useful for motivating how far robotics as a field has to go in order to succeed in the real world. Examining examples of robot failures is an important exercise to see how robot fails, what conditions cause failure, and what the field can work on in order to make robots better.  Segmenting out failure cases and discussing what issues occurred during trials gives the viewer a more holistic understanding of how robust and reliable the system is.

\citet{pan2017agile} record raw footage of an autonomous car with an onboard camera driving around multiple times around a track outside (video at \cite{record1}). Simply showing long raw footage of many trials back-to-back from either a continuous take or clear film edits, as outlined in section \ref{sec:clearfilmedits}, is a great way to demonstrate how robust and reliable the system is compared to just showing one trial.

We offer an example video made by the collaborators of this paper's authors that does not accurately reflect the robustness of the robot system (video at \cite{badrecord1}). Here, the authors demonstrate the capability of their Virtual Reality teleoperation system to be used in order to perform a long-distance complex cup stacking task. While the video includes a single continuous shot of a user successfully performing the task one time, it does not show other attempts or indicate how many attempt were required to get the final success trial. When a video shows only the successful trials, it should indicate whether it was done in a single attempt or how many trials were needed to get the ones in the videos. A robot video should convey to the viewer an accurate sense of how robust and reliable the robot system is.

\subsection{Display Film Rate}
Robot videos are often sped up or slowed down in order to help the viewer better understand what is going on. This is because robots can be fast and difficult to see in real-time, or robots can be slow in their processes and a shorter video is desirable. While manipulating film rate is a useful technique for editing robot videos, it is important to clearly indicate the film rate (2x speed, $\frac{1}{2}$x speed, etc.) so the user has an accurate sense of how fast the robot operates. This is perhaps the most obvious to include maxim, since displaying the film rate is so common (video examples at \cite{rate1} and \cite{rate3}).

\citet{mirjan2016building} employ an interesting technique of displaying their drone build a bridge at both a normal speed and a sped up speed (video at \cite{rate2}). The video first shows the drones operating at a regular pace, and then speed up the video to 20x, and then slow down the video back to real time at the end. Showing the real-time and manipulated time back-to-back help the viewer get a sense of how long robot would takes to operate.

\subsection{Highlight Supporting Systems}
Robot videos often include many supporting systems such as AR tags, speech recognition systems, motion capture, and wizard-of-oz teleoperation and human scripted behavior. Supporting systems are useful for demonstrating how the research contribution can be integrated into a full end-to-end robot system to perform a truly complex and impressive task. When using supporting systems in a robot video, it is important to indicate what the supporting systems are. Supporting systems have their own unique limitations, (for example, motion capture requiring a structured setup), and viewers should have an accurate and complete understanding of the robot's capabilities and supporting systems in the video.

\cite{support1} is a positive video example that explicitly highlights the supporting systems that are used. They point out where their pressure regulators and external cameras are and how they are used for the movement of the soft robot. \cite{support2} is also a good video example that visually points out how their robot arm uses supporting systems for pick and place by focusing on the cameras and QR tags that the robot uses for scanning objects. Although certain systems in a robot video may be easily identifiable by experts, one should not assume all viewers are able to identify and understand the context in which subsystems are being used, and the video should be explicit and explain, to the best of their reasonable capability, what the supporting systems are, where they are, and how they are being used.

We offer a robot video example made by the authors of this paper that do not make the supporting systems explicitly clear. In \citet{karamcheti2017tale}, a robot is following natural language instructions by using a motion capture as a supporting system to know where the blocks and the robot are localized (video at \cite{badsupport}). However, the video does not make this explicit, and it is difficult to tell whether the robot is perceiving the block using its on-board sensing equipment or not. Robot videos should highlight supporting off-the-shelf syetems and be clear about how they are being used with the research contribution.

\section{Checklist}
In order to help put our maxims to practice, we offer a checklist that can be used when producing robot videos:
\begin{itemize}
  \item Record All Trials
  \begin{todolist}
    \item Keep all recordings and upload them along with a shorter video (if possible).
    \item Segment out failure recordings to highlight where system goes wrong.
    \item Indicate how many trials it took to get the successful one shown in video.
  \end{todolist}
   \item Clear Film Edits
  \begin{todolist}
    \item Indicate when spliced shots are from different trials (e.g: film edits are not from one continuous take).
    \item Make clear where various camera views are being captured from.
  \end{todolist}
  \item Display Film Rate
  \begin{todolist}
    \item If you manipulated speed of video, display the film rate.
  \end{todolist}
  \item Highlight Supporting Systems
  \begin{todolist}
    \item Whenever using off-the-shelf supporting systems to demonstrate your end-to-end system (QR tags, motion planners, speech-to-text, human scripted behavior, etc.), make clear what they are and where they are and how they are being used with your contribution.
    \item Show what they are, where they are, and how they are being used.
  \end{todolist}
\end{itemize}

\section{Related Work}
Simulators make robot research much more easily reproducible and verifiable than needing to use real robot hardware, and we refer the interested reader to \cite{islam2017reproducibility} and \cite{2020arXiv200312206P} for more in-depth analysis of the subject of reproducible research for learning agents with simulators. Corpus based evaluations are also useful for standard comparison of real-robot systems, with standard datasets being useful in a variety of robotic subfields \cite{calli2017yale,nardi2015introducing,moll2015benchmarking}. All of these standard methods of validating robot research should be done if possible when conducting robot research. We also recognize that there are many types of robot research, both systems papers and theoretical papers, that have cases where a video on a real robot is not necessary or practical.

\section{Conclusion}
In this article, we offer a set of best practices when making robot videos: \textbf{make clear film edits}, \text{reord all trials}, \textbf{display film rate}, and \textbf{highlight supporting systems}. We hope that our maxims spark a discussion in the robotics community about the value of robot videos and that our examples provide helpful context for our video suggestions. We look forward to engaging with the community and discussing how to take further steps towards developing best practices for robot videos.

\section{Acknowledgements}
We would like to thank Jessica Forde for reviewing and giving advice on the checklist. 
\bibliographystyle{plainnat}
\bibliography{bibliography}

\end{document}